\documentclass[conference]{IEEEtran}
\IEEEoverridecommandlockouts
\usepackage{cite}
\usepackage{amsmath,amssymb,amsfonts}
\usepackage{algorithmic}
\usepackage{graphicx}
\usepackage{textcomp}
\usepackage[table]{xcolor}
\usepackage{subcaption}
\usepackage[justification=justified]{caption}
\usepackage{multirow}
\def\BibTeX{{\rm B\kern-.05em{\sc i\kern-.025em b}\kern-.08em
    T\kern-.1667em\lower.7ex\hbox{E}\kern-.125emX}}
    
\begin{document}

\title{Enhancing Classification with Hierarchical Scalable Query on Fusion Transformer}

\author{\IEEEauthorblockN{Sudeep Kumar Sahoo}
\IEEEauthorblockA{\textit{Samsung R\&D Institute} \\
Bangalore, India \\
sudeep.sahoo@samsung.com
}
\and
\IEEEauthorblockN{Sathish Chalasani}
\IEEEauthorblockA{\textit{Samsung R\&D Institute} \\
Bangalore, India \\
sathish.c@samsung.com
}
\and
\IEEEauthorblockN{Abhishek Joshi}

\IEEEauthorblockA{\textit{Samsung R\&D Institute} \\
Bangalore, India \\
abhi.joshi@samsung.com
}
\and
\IEEEauthorblockN{Kiran Nanjunda Iyer}
\IEEEauthorblockA{\textit{Samsung R\&D Institute} \\
Bangalore, India \\
kiran.iyer@samsung.com
}
}
\maketitle

\begin{abstract}
Real-world vision based applications require fine-grained classification for various applications of interest like e-commerce, mobile applications, warehouse management, etc. where reducing the severity of mistakes and improving the classification accuracy is of utmost importance. This paper proposes a method to boost fine-grained classification through a hierarchical approach via learnable independent query embeddings. This is achieved through a classification network that uses coarse class predictions to improve the fine class accuracy in a stage-wise sequential manner. We exploit the idea of hierarchy to learn query embeddings that are scalable across all levels, thus making this a relevant approach even for extreme classification where we have a large number of classes. The query is initialized with a weighted Eigen image calculated from training samples to best represent and capture the variance of the object. We introduce transformer blocks to fuse intermediate layers at which query attention happens to enhance the spatial representation of feature maps at different scales. This multi-scale fusion helps improve the accuracy of small-size objects. We propose a two-fold approach for the unique representation of learnable queries. First, at each hierarchical level, we leverage cluster based loss that ensures maximum separation between inter-class query embeddings and helps learn a better (query) representation in higher dimensional spaces. Second, we fuse coarse level queries with finer level queries weighted by a learned scale factor. We additionally introduce a novel block called Cross Attention on Multi-level queries with Prior (CAMP) Block that helps reduce error propagation from coarse level to finer level, which is a common problem in all hierarchical classifiers. Our method is able to outperform the existing methods with an improvement of about 11\% at the fine-grained classification. 

\end{abstract}

\section{Introduction}

Fine grained classification is of paramount importance in medical applications like recognizing plant diseases accurately, human skin conditions, etc. It can also enhance user experience in applications like e-commerce with better search and retrieval by accurately recognizing and differentiating fine-grained objects. We propose improving fine grained classification through a hierarchical classification approach. Hierarchical classification is a way to group things according to a hierarchy or levels, thereby reducing the classification tasks into multiple sub-tasks to reduce the difficulty of the original task. Classification methods often assume a flat label hierarchy. However, most real-world data have relations between the categories, which can be exploited through multi-level hierarchies to improve accuracy. Using hierarchy to guide the classification models is very similar to the way humans deal with visual understanding. Bringing hierarchy into the classifier models helps encode rich correlations among the categories across different hierarchy levels, thereby reducing inter-class confusion. Models trained in a hierarchical fashion and inferencing level by level in classification improve the explainability and interpretability of image understanding models. 

\begin{figure}[t]
\centerline{\includegraphics[width=0.5\textwidth]{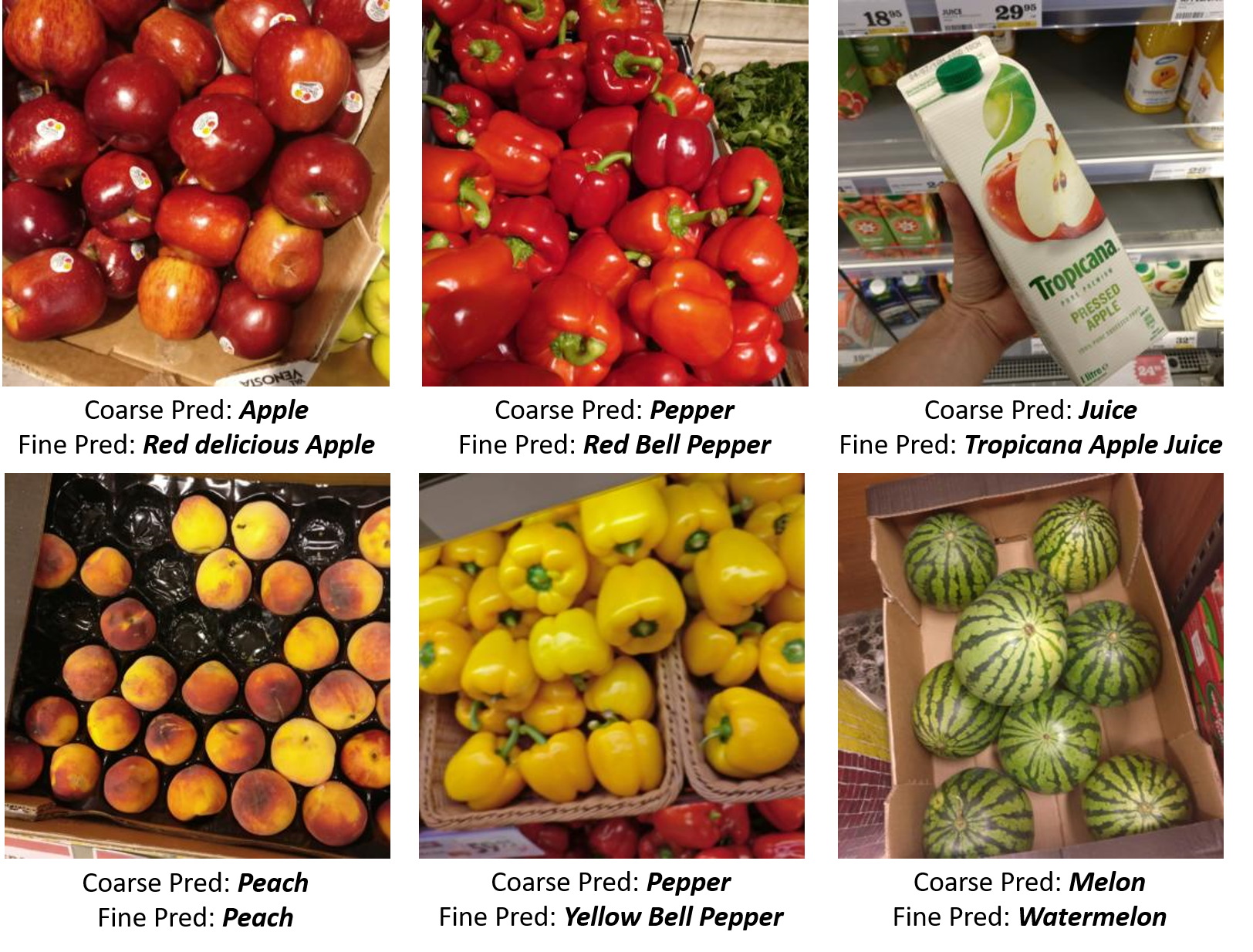}}
\caption{Some example images from GroceryStore dataset\hspace{-1mm}~\cite{klasson2019hierarchical} and our predictions for them. The model is able to distinguish and correctly categorize between similar looking images}
\vspace{-3mm}
\label{fig:example1}
\end{figure}

So far, attempts have been made to achieve hierarchical classification tasks with Convolution Neural Networks(CNN). Convolutional Neural Networks have remained dominant and have largely been the architecture of choice for computer vision tasks. Intermediate feature map representations learned by CNNs in classification tasks have proven to be powerful descriptors. Given their extensive use CNNs have been the core integral component in building networks aimed at almost all vision tasks, given their ability to model 2D local structures and their inherent properties of invariance and equivariance. Although CNN has a strong influence in vision related tasks, they suffer from a lack of robustness, which leads to wrong predictions in surprisingly easy scenarios. \cite{wang2022can} explore more on a lack of robustness in CNN architectures, in contrast to transformer based architecture. Transformers \cite{vaswani2017attention, devlin2018bert} on the other hand have become very prominent in Natural Language Processing. Recent advances in the application of transformers in language tasks have led researchers to explore attention based transformers for computer vision \cite{dosovitskiy2020image}. However, training Transformer architectures require large amounts of training data due to the absence of strong inductive biases. To get the best of both CNNs and Transformers, Convolutions are introduced \cite{wu2021cvt} to ViT structure to improve performance and robustness. Inspired by the design of CvT we introduce transformer blocks on Convolutional Neural Network layers in our hierarchical fusion modules, to get the best of both worlds.

Hierarchical classification suffers from error propagation. The error caused at a certain level will directly impact the performance of its next level. In natural images, objects sometimes might need to be localized in context, most often due to their small size in images, occlusion scenarios, inter-class confusions, etc to get the right predictions. Query based classification approaches have proved to be effective in probing the image for the presence of objects. To this end, we propose query based hierarchical image classification with CNN as a backbone. We design a query approach that is scalable across multiple levels, supporting a large number of classes. For the query to have the best initial representation, we initialize the query with a weighted eigen image, which we call Eigen-Query. Eigen images in general span an optimal linear subspace. However, it has maximum scatter within the linear subspace over the entire imageset. The ideal way of query representation is to have maximum between-class scatter(i.e low cosine similarity between a class query and inter-class samples) while having minimum within-class scatter(i.e high cosine similarity between the class query and its class samples), similar to Fisher Images. In order to transform Eigen-Query to Fisher-Query, we use Cluster focal loss, which explicitly maximizes between-class scatter and minimizes within-class scatter for all Eigen-Queries at every hierarchical level. To provide scale invariance, we introduce fusion transformers that fuse high-resolution low-level semantics with low-resolution high-level semantic feature maps to maximize object size invariance.

\begin{figure*}
\centerline{\includegraphics[width=\textwidth]{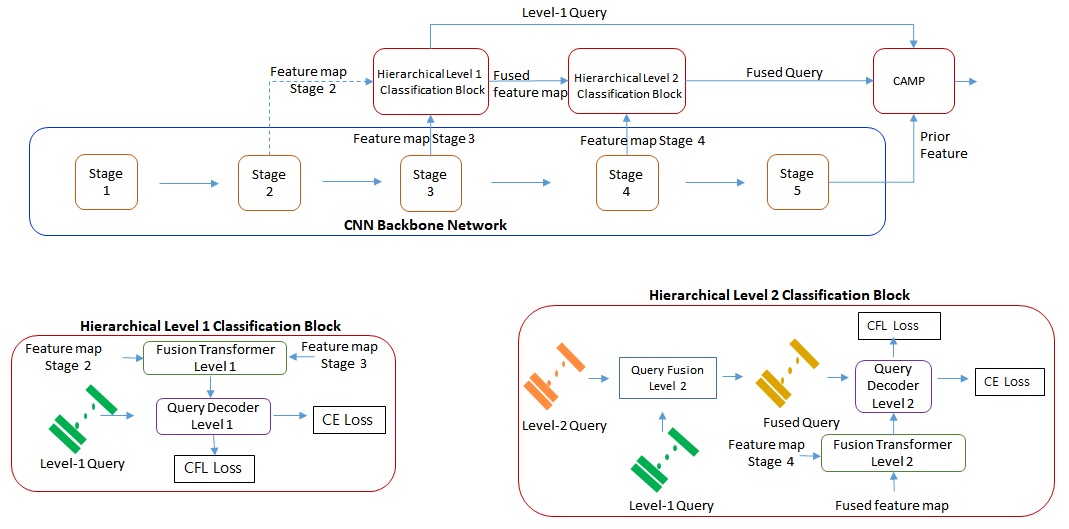}}
\caption{Proposed architecture block diagram. We denote Level 1 as Coarse-level and Level 2 as Fine-level Hierarchy.}
\vspace{-3mm}
\label{fig:block_diag}
\end{figure*}

The main contributions of this paper are as follows:
\begin{itemize}
    \item Scalable learned Fisher-Query based attention module, that comprises coarse to finer query fusion for better representation.
    \item Fusion transformer blocks for multi-scale representation. It enhances the accuracy of small-size objects.
    \item Cross Attention on Multi-level queries with Prior (CAMP Module) that reduces error propagation and improves propagating higher coarse level accuracy to finer levels.
\end{itemize}

\section{Related Work}

There have been rich studies on exploiting hierarchies to improve accuracy in computer vision and natural language processing (NLP) domains. Several techniques have been employed in Computer Vision tasks, that have previously been proposed in NLP related domains. We discuss the prior works in Hierarchical Classification for each domain as follows:

\subsubsection{Hierarchical Classification in Computer Vision}
In \cite{zhou2010multi}, with the model being hierarchy aware, authors propose weighted multi-class logistic regression formulation by considering the consistency of cost before rescaling for multi-class cost-sensitive learning. Similarly, \cite{wu2021cvt} proposed a multi-level optimization framework using class re-weighting and combining losses at different hierarchies of the tree. \cite{he2021hierarchical} proposes hierarchical image classification with Unsupervised Domain Adaptation(UDA) technology by fusing features learned from a hierarchy of labels. The fused features are used to predict the finest-grained class. Interestingly \cite{dhall2020hierarchical} use label-hierarchy knowledge to improve on visual semantics of images. Further, they explicitly model label-label and label-image interactions to guide hierarchical image classification and representation learning. \cite{karthik2021no, wu2016learning} have tried to exploit label hierarchy and graph distances to reduce mistake severity in hierarchy aware deep image classifiers. \cite{chen2018fine} exploit Hierarchical Semantic Embedding(HSE) to sequentially predict category score vector at each level in the hierarchy from coarse to fine level. The predicted score vector of coarser level is used as prior knowledge to learn finer-grained feature representation.

\subsubsection{Hierarchical Classification in Natural Language Processing}

Recently, the graph-based text classification method has been gaining attention that transforms the text classification task into a graph classification task or a node classification task and has shown to be promising.  For the learning of graph representation, various GNN models are used in text classification, such as GCN~\cite{Yao_Mao_Luo_2019, Liu_You_Zhang_Wu_Lv_2020}, GNN~\cite{zhang-etal-2020-every}, and GAT-based model~\cite{linmei-etal-2019-heterogeneous}. Le \textit{et al.} \cite{InferringConceptHierarchies} focus on the task of inferring relationships from large text corpora. Different from other methods, this work proposes the hierarchical nature of hyperbolic space that allows to learn highly efficient representations and to improve the taxonomic consistency of the inferred hierarchies. Prior work by Morin \textit{et al.} \cite{pmlr-vR5-morin05a} introduced a hierarchical decomposition of the conditional probabilities that yields a speed-up during both training and recognition. The hierarchical decomposition is a binary hierarchical clustering constrained by the prior knowledge extracted from the WordNet semantic hierarchy.

\subsubsection{Query based approaches in vision} Studies on query based approaches have been done to accuracy on several tasks. \cite{liu2021query2label} demonstrates an effective approach to solving the multi-label classification problem by querying the existence of class labels through transformer decoders. However, the approach fails to scale across extreme classification tasks. To overcome this problem \cite{ridnik2021ml} proposed a new attention based classification head to query the existence of labels through a group-decoding scheme that makes querying highly effective and scalable across many classes. Query based approaches have also been effectively used in image captioning. \cite{fang2022injecting} proposed to use transformer based image captioning model with Concept Token Network(CTN) that predicts concept tokens, which are used to generate end-to-end captioning.

\section{Proposed Method}

The core idea behind transformer based decoder is to retain the rich information from the convolutional feature map which was discarded in the conventional global average-pooling method. Also, the multi-head attention mechanism decouples object representations into multiple parts for better performance and interpretability. However, the traditional query based transformer decoder suffers from scalability issues with an increasing number of classes and levels. To overcome these shortcomings, we propose a novel dynamic query based approach with a limited number of dynamic information-rich queries at each stage of classification.

\subsection{Architecture Design}

Initially, the input image is passed through an existing pre-trained model as backbone to obtain convolutional feature representation. In our experiment, we have used Densenet169~\cite{huang2017densely} as our backbone to extract features for the decoder. But our method is backbone-agnostic i.e., the fusion transformer block(FT Block) can be plugged with any feature extractor. The features were taken from different layers of the backbone to accommodate diverse semantics and spatial information across multiple stages of the hierarchical classification. The backbone is further tuned at a lower learning rate to make the features specific to the input dataset. The feature channels are projected to a fixed and desired query dimension before passing through the FT blocks for consequent stages. Both sets of feature maps at coarse and fine level are passed through similar decoder blocks and done in a sequential manner to absorb class hierarchy knowledge into the model.

At each level of granularity, multiple consecutive layers of representations are chosen to be fused at an interim spatial dimension. The fusion of the layers is done using a CvT based convolutional transformer module with multi-head cross attention. The low-resolution high semantics layer is supplied as the query and higher resolution features are passed as key/value with positional embeddings. The computed cross attention is able to highlight semantic details in a high-resolution feature map. The layers were merged progressively from the backbone as shown in Fig. \ref{fig:block_diag} i.e., the fusion output from one level is passed on to the next along with the next higher semantic feature from the backbone for merging. This way of succession fusion incorporates rich semantic and spatial information at later stages in the hierarchical classification.

Fused representation at each level is passed through a Convolutional transformer based decoder. The label embeddings were fed as queries and the image representation as key/value to compute multi-head cross attention. The queries passed were treated as learnable parameters in end-to-end training to model label correlations implicitly. The output of the transformer is passed through a couple of convolution blocks with ReLU activation function. This decoded output is finally passed through a global average pooling(GAP) layer to generate the output logits.

\subsection{Scalable Query}

Instead of assigning each category a single new query, our approach has the capability to represent multiple categories under each query. At each level, the maximum number of queries is limited to the highest no. of possible subclasses for a fixed class at the preceding level. The query embeddings from all prior predicted superclasses are merged with the current query in a weighted manner to dynamically encapsulate the present class information. Due to the sequential query fusion, the embeddings at fine level are class specific and have clear semantic meanings associated. The fused query $Q_2$ at fine level is implemented as the following equation:

\begin{equation}
Q_2 = [\beta*Q'_2+(1-\beta)*Q^p_{1,l1}]*mask
\label{eq:query}
\end{equation} 

where $\beta \in [0, 1]$ is a trainable weighting parameter used to fuse top-level query with the current level. $Q'_2$ is the base label embeddings at fine level. $Q^p_{1,l1}$ is the projected embedding for level 1 class(coarse) to which the image belongs. During training, embedding for ground truth at coarse level is used to create a weighted query to minimize error propagation through the hierarchy while during inference, predicted coarse label is used. Masking is used at fine granularity to handle the variation in the number of subclasses in hierarchy distribution.

This approach of query designing reduces the decoder operation and time complexity by a factor of $N_{i}/k_{i}$ where $N_i$ is the number of classes at $i^{th}$ level and $k_i$ is the maximum no. of queries permissible at that granularity. The dynamic query composition ensures that the model parameters and time complexity are independent of the number of classes and proportional to the maximum permissible queries at any hierarchy. Hence, with the addition of new classes at finer levels, the model parameters won’t increase linearly. This makes the architecture scalable for larger datasets and generalized to continuous expansion of the input data. The model is trained in an end-to-end manner while incorporating and exploiting the previous class information at each stage via dynamic queries.

\subsection{Query Initialization}

The label queries are designed as 2D feature maps which capture the essence of images from each class in a reduced dimension. To create a meaningful representation, they are initialized using principal component analysis(PCA) as described below. First, the images were normalized and covariance matrix for each fine class is calculated. Then we extract the eigenvectors of these matrices to determine the principal components and compute the class query maps by taking an eigenvalue based weighted combination of these vectors. The coarse and fine level Eigen-Queries were created from these maps based on their respective leaf class associations. 

The initial queries, however, suffer from maximum scatter across the linear eigensubspace over the dataset. The ideal query representations should have maximum separation between classes like Fisher's discriminant. Also, they should help the same class image features form a compact cluster. The usage of Cluster Focal Loss, mentioned in subsection \ref{subsect:CFL}, ensures these conditions are met to transform the learnable Eigen-Queries into Fisher-Queries. 

\subsection{Cluster Focal Loss}
\label{subsect:CFL}

Motivation for Cluster Focal Loss\cite{sem_energy_ijcnn2022} is to get maximum separation between query embeddings synonymous with class cluster center. Learning better class-wise cluster representation depends on two key factors, the need to mitigate the ill effects of large class imbalance and the need to differentiate between hard and easy examples. We employed a simple method to tackle this by calculating semantic similarity of query w.r.t. average pooled intermediate feature map. 

\begin{figure}[!ht]
\centerline{\includegraphics[width=0.5\textwidth]{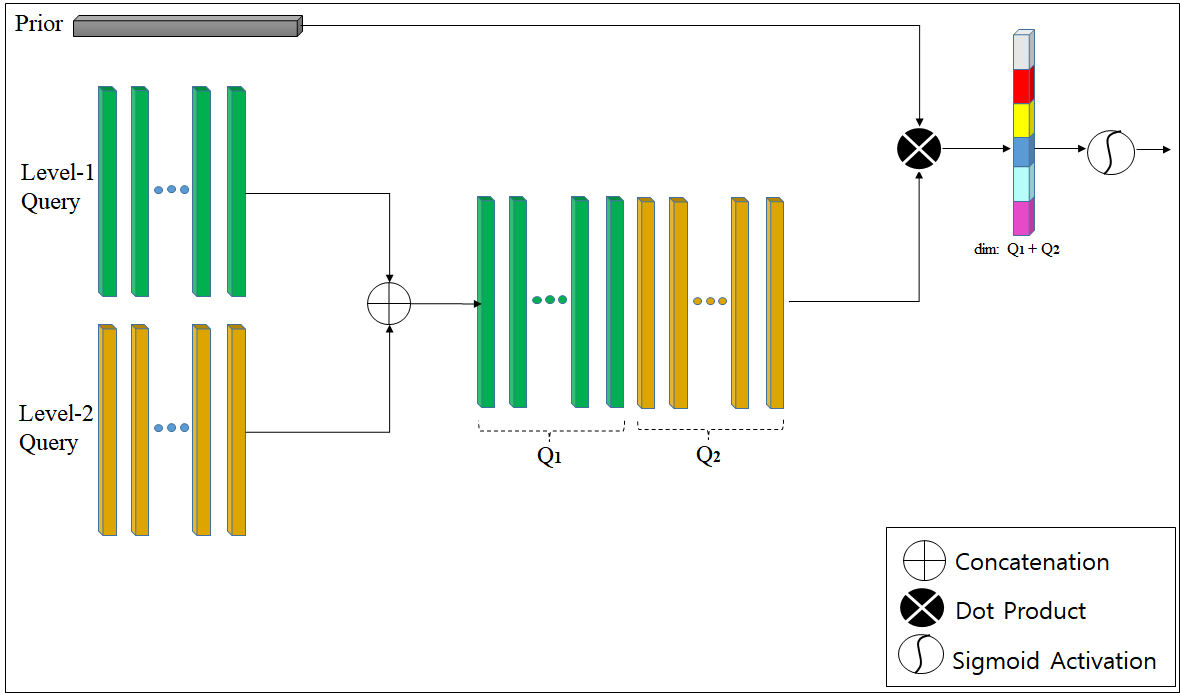}}
\caption{Cross Attention on Multi-level queries with Prior (CAMP) Block architecture.}
\label{fig:CAMP_block}
\end{figure}
\vspace{-5mm}

\subsection{Cross Attention on Multi-level queries with Prior (CAMP)}
We introduce a novel mechanism to improve the error propagation induced by the class logits by providing cross attention on multi-level query embedding and an implicit prior. The backbone network's penultimate layer output is rich in semantic information and can be suitably leveraged to serve as prior information for error correction. This proposed block is referred to as CAMP Block and is illustrated in Figure \ref{fig:CAMP_block}. This module can be easily plugged into the existing architecture and can be trained in an end-to-end manner.

As shown in the figure, Multi-level queries are fused using concatenation operation. To this joint representation, cross attention is applied with the Prior feature vector through a dot product. The final output vector obtained is of size $Q_1 + Q_2$, where $Q_1$ and $Q_2$ are the number of queries at Level-1 and Level-2 respectively. We apply Sigmoid activation function to the final output and train with Binary Cross Entropy Loss.

\section{Experiments and Results}

\begin{table}[ht]
\centering
\rowcolors{3}{}{gray!20}
\begin{tabular}{c | c c}
    \hline
    \multirow{2}{5em}{\textbf{METHOD}} & \textbf{COARSE} & \textbf{FINE} \\
    & \textbf{ACCURACY(\texttt{\%})} & \textbf{ACCURACY(\texttt{\%})} \\
    \hline
    Densenet-scratch\cite{huang2017densely} & 75.67 & 67.33 \\
    \hline
    Softmax & 83.34 & 71.67 \\
    \hline
    AE + Softmax & 82.42 & 70.69 \\
    \hline
    VAE + Softmax & 81.24 & 69.20 \\
    \hline
    VCCA\cite{klasson2020using} & 82.12 & 70.72 \\
    \hline
    Ours & \textbf{88.43} & \textbf{81.33} \\
    \hline
\end{tabular}
\caption{The column Coarse Accuracy corresponds to the high-level classification accuracy. The column Fine Accuracy corresponds to classification within the correct parent class.}
\label{tab:accuracy}
\end{table}
\vspace{-2mm}

\subsubsection{Dataset} Klasson et al. \cite{klasson2019hierarchical} collected images from fruit, vegetable, and refrigerated sections with dairy and juice products in 20 different grocery stores. The dataset consists of 5,421 images from 81 different classes. The dataset is hierarchical in nature, having two levels consisting of coarse and fine labels. The total number of classes in first and second levels are 43 and 81 respectively. A coarse-grained class without any further subdivision in the dataset is considered a fine-grained class in the classification experiments.

\begin{figure}[ht]
\centerline{\includegraphics[width=0.5\textwidth]{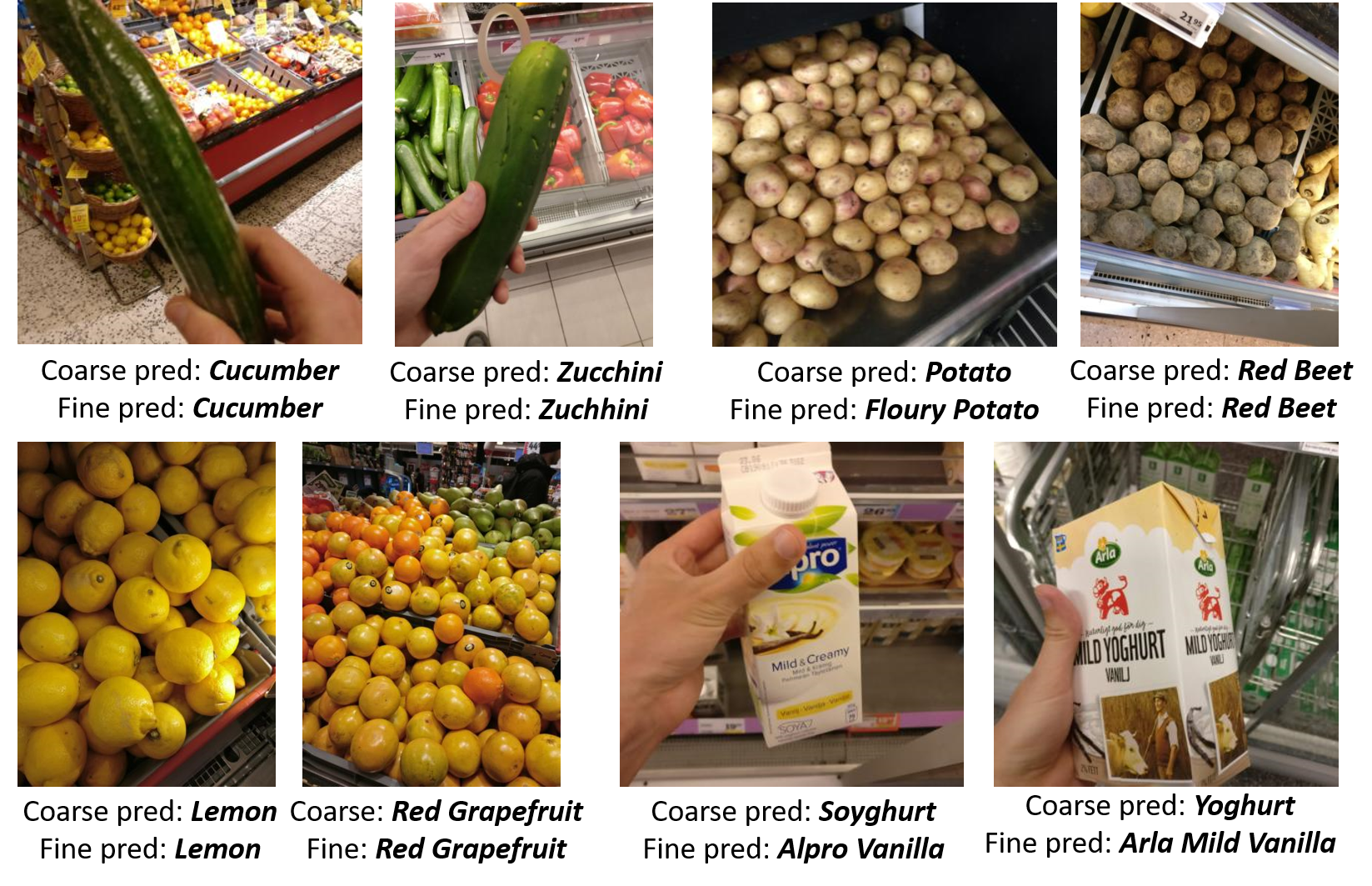}}
\caption{Correctly predicted sample test set images at both coarse and fine level}
\vspace{-1mm}
\label{fig:res_correct}
\end{figure}

\subsubsection{Qualitative Results} 
\begin{figure}[ht]
\begin{subfigure}{0.48\textwidth}
\includegraphics[width=1.0\linewidth]{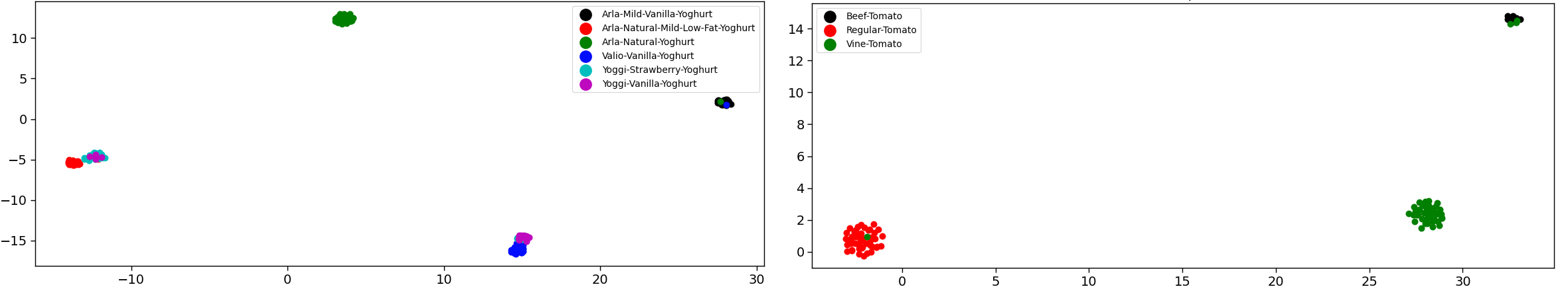}
\end{subfigure}
\caption{UMAP plots for our trained Fisher-Queries on Yoghurt and Tomato categories at the fine-level}
\label{fig:UMAP_Fisher}
\end{figure}

\begin{figure}[ht]
\begin{subfigure}{0.48\textwidth}
\includegraphics[width=0.98\linewidth]{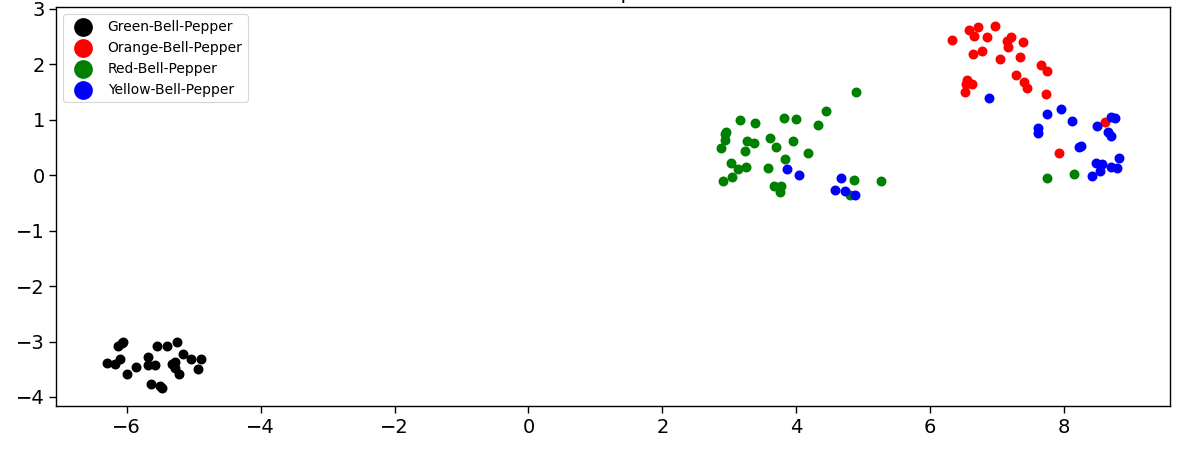}
\end{subfigure}

\begin{subfigure}{0.48\textwidth}
\includegraphics[width=0.98\linewidth]{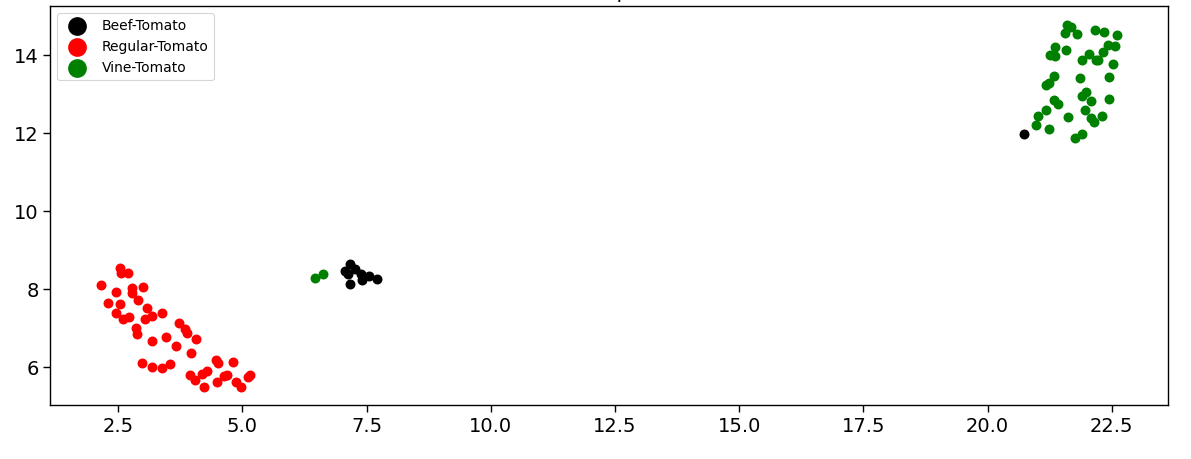}
\end{subfigure}

\caption{UMAP plots for our trained model embeddings on Pepper and Tomato categories at the fine-level respectively.}
\label{fig:tSNE}
\end{figure}

In Fig. \ref{fig:UMAP_Fisher} we present the learned Fisher-Queries representations at Fine-level through UMAP\cite{sainburg2021parametric}. In Fig. \ref{fig:tSNE}, we compare UMAP representations of learned logits for our trained model. For instance, the top figure in the plot in Fig. \ref{fig:tSNE} represents class clusters representation of the class Pepper having 4 sub-classes. We observe relatively compact clusters, which can be attributed to training with CFL loss which respects class-wise cluster centers. 

\begin{figure}[h]
\centerline{\includegraphics[width=0.5\textwidth]{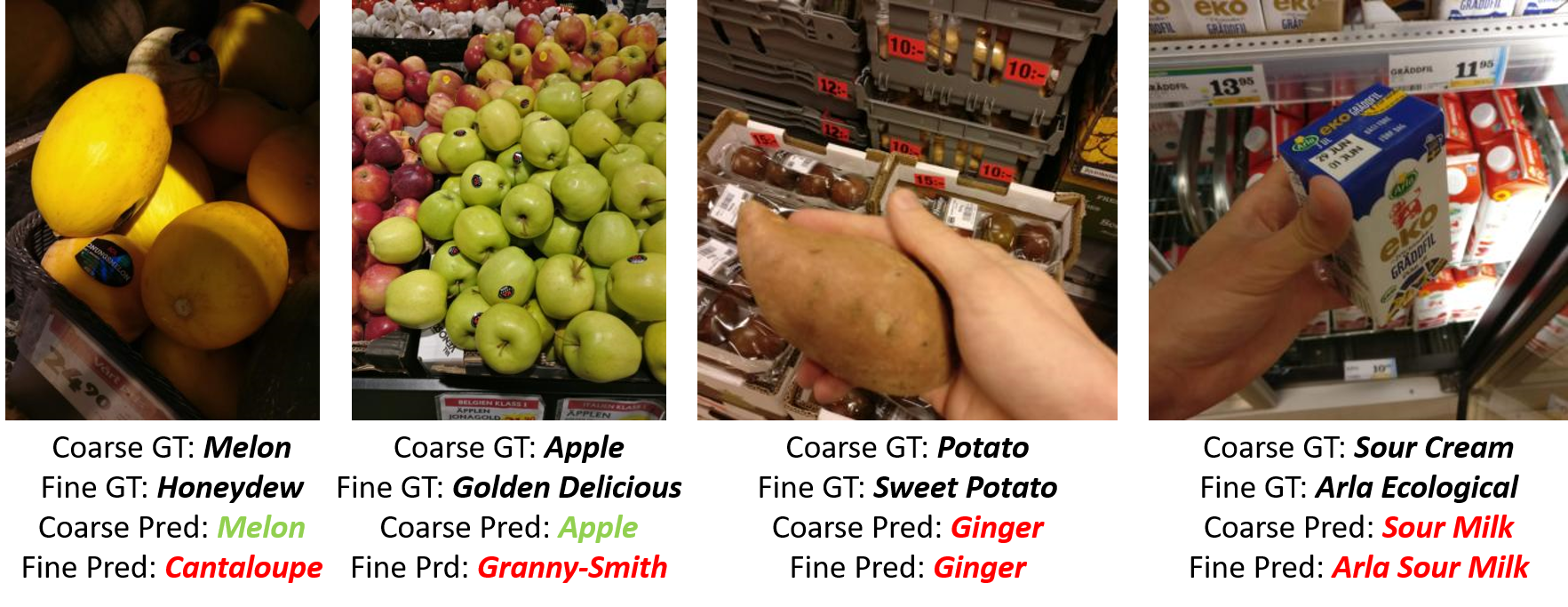}}
\caption{Coarse and fine level predictions for sample test images where our system predicts wrongly }
\label{fig:res_wrong}
\vspace{-5mm}
\end{figure}

We present some visually similar images from the test set which are correctly predicted by our system in Figure \ref{fig:res_correct}. From the results, we can observe that our model can distinguish between very close-looking images from different categories. We also provide some of the incorrectly predicted samples in Figure \ref{fig:res_wrong}. From our observations, some of the major misclassifications at coarse level occurred in visually-resembling fruit categories like apple-peach, lemon-grapefruit, satsumas-orange, kiwi-passionfruit, etc. Furthermore, the fine level classifications are hard to differentiate for sub-categories under apple, melon, pear, potato and pepper classes. The model can rightly classify fine-grained classes in hard images, however, for visually very challenging scenarios the predictions are at times incorrect. From the examples of Figure \ref{fig:res_wrong}, we can see that images containing items from multiple categories, partly occluded images and images which are tough to distinguish even by humans are mispredicted by our system occasionally.

\subsubsection{Quantitative Results}
In Table \ref{tab:accuracy}, we showcase quantitative results. We benchmark both the coarse level accuracy and the fine-grained classification accuracy. Prior methods such as \textit{Densenet scratch} refer to DenseNet169 architecture trained from scratch, \textit{Softmax} refer to softmax classifier trained on off-the-shell features, \textit{AE} is the abbreviation for Auto Encoder and \textit{VAE} is the abbreviation for Variational Auto Encoder. for  We observe that our method achieves state-of-the-art for both Coarse and Fine accuracy indicating that our approach is promising and scalable. Our method works significantly better on fine level classification since our approach of dynamic query encapsulates top-level hierarchy information to improve lower levels. Overall, we achieve a good improvement in performance by about 5\% and 10\% in Coarse and Fine-level classification respectively. The use of scalable dynamic query modeling and query fusion from coarse to fine granularity helped the system improve the fine level accuracy greatly.

\vspace{-3mm}

\vspace{-3mm}

\section{Ablation Analysis}
To further understand the contribution of each component to the performance of the proposed model, we perform an ablation analysis. We incrementally add each module to the setup. We start with the scalable query fusion model and refer to it as \textit{Base} setup. Next, we incrementally add CFL loss component. Finally, we add CAMP block to the setup. From Table \ref{table:ablation}, it can be observed that CFL helps in improving fine-level accuracy by about 1\% over the base method. This highlights the effectiveness of including CFL loss to learn better and well-separated inter-class query embeddings at fine level. Thirdly, we add Fisher-Queries initialization to it and observe some improvement. Next, when CAMP block is added to the experimental setup there is a further boost of 1\% Fine accuracy. The improvement in accuracy can be attributed to the cross attention mechanism of multi-level queries with the prior feature vector that helps in reducing the error propagation and multi-level predictions learning to be correct simultaneously across every level. Finally, we include Fisher-Queries initialization on the previous setup and results in best performing model. Overall, we observe about 4\% improvement in Fine accuracy against the Base method. Thus, the ablation study justifies the efficacy of each of the novel contributions to the overall performance.

\begin{table}[h]
\begin{center}
\begin{tabular}{l|c|c}
\hline
    \multirow{2}{5em}{\textbf{METHOD}} & \textbf{COARSE} & \textbf{FINE} \\
    & \textbf{ACCURACY(\texttt{\%})} & \textbf{ACCURACY(\texttt{\%})} \\
    \hline \hline
    Base & 86.29 & 77.02  \\
    Base + CFL & 86.33 & 78.18 \\
    Base + CFL + Eigen & 86.89 & 79.31 \\
    Base + CFL + CAMP & 87.58 & 80.97 \\
    Base + CFL + CAMP + Eigen & \textbf{88.43} & \textbf{81.33} \\
    
                    \hline
\end{tabular}
\end{center}
\caption{Ablation Results of our proposed approaches. Bold represents superior results.}
\label{table:ablation}
\end{table}
\vspace{-5mm}

\section{Conclusion}

In this paper, we proposed a novel and effective approach that hierarchically learns scalable query embeddings to improve fine-grained classification achieving state-of-the-art performance. Implicitly, the scalable query based approach makes it suitable for extreme classification problems with several hierarchy levels. We adopt cluster based loss function to improve on modeling query embedding and learn a better representation of class-wise cluster centers. To mitigate the problem of error propagation from coarse level to fine level classification, we add cross attention mechanism(CAMP Block) for multi-level queries to further enhance the accuracy. Our future work will involve, demonstrating the effectiveness of our approach on extreme classification datasets and adapting it for weakly supervised settings.

{\small
\bibliographystyle{ieee_fullname}
\bibliography{egbib}
}

\end{document}